# A tree augmented naive Bayesian network experiment for breast cancer prediction

Ping Ren

## Abstract

In order to investigate the breast cancer prediction problem on the aging population with the grades of DCIS, we conduct a tree augmented naive Bayesian network experiment trained and tested on a large clinical dataset including consecutive diagnostic mammography examinations, consequent biopsy outcomes and related cancer registry records in the population of women across all ages. Our tasks are to classify the conventional "Benign vs. Malignant" and the new "Benign/LG vs. IntG/HG/Invasive" based on mammography examination features and patient demographic information, specifically to predict the probability of malignancy, for the biopsy threshold setting and the biopsy decision making. The aggregated results of our ten-fold cross validation method recommend a biopsy threshold higher than 2% for the aging population. The Receiver Operating Characteristic curves and the Precision-Recall curves by aggregating the ten-fold cross validation results are interesting.

## Introduction

The practice of mammography in aging populations successfully diagnoses breast diseases including invasive cancers at an earlier stage. [1-2] However, the efficacy of mammography in older age groups can be affected substantially by inherent problems such as false-positive [3-4] and over-diagnosis. [5] The false-positive problem leads to the problem of the high rate of breast biopsy: in the population of U.S. women older than 65 there are 140,000 breast biopsies cases per year, [6-7] 75% of which turn out to be benign findings. Note that the procedure of breast biopsy is the most expensive component of breast cancer diagnosis. [8] These years the problem of the high rate of breast biopsy becomes more and more urgent: there are currently 21,784,000 women over age 65 and the first of the baby boom generation born in the year 1946 has been aged over 65 since the year 2011; it is also projected that the number of women over age 65 will double and the number of women over age 85 will increase fivefold from the year 2000 to the year 2050. [9]

Another urgent problem emerges from the increasing rate of DCIS (ductal carcinoma *in situ*) which is a non-obligate precursor to subsequent invasive breast cancer. [10-11] DCIS, on one hand, typically appears as microcalcifications on mammography, whereas microcalcifications could be related to benign conditions including fibrocystic changes, a fibroadenoma, or fat necrosis. [12] The microcalcifications are often pursued with biopsy for diagnosis, which leads to a low positive predictive value of biopsy. As a result, the 2009 National Institutes of Health (NIH) consensus conference on DCIS urges the development of methodologies to more accurately identify subsets of women who might not need the treatment for DCIS [13] and whose risk of progression could be low enough to employ watchful waiting (mammographic evaluation at short term intervals) rather than breast biopsy. [14-15]

On the other hand, DCIS may remain indolent for sufficiently long that a woman dies of other causes. [16-17] Progression from DCIS to invasive breast cancer can be predicted by grades. [10,18-19] Pathologists use three grading categories: grade 1 or “low grade” (LG), grade 2 or “intermediate grade” (IntG), and grade 3 or “high grade” (HG). [20] Study suggests that patients with DCIS of any grade are at increased risk for developing breast cancer, among which the interval is longest for the low grade. [10,17] Age adjusted incidence rate of DCIS between 1973 and 2000 increased from 4.3 to 32.7 per 100,000 women-years, equivalent to an increase of 660%, [21] the majority of which were detected on mammographic screening. [22] The increased rate of DCIS was most notable in the group of women over age 50. [23] Consequently, the NIH statement on DCIS highlighted the need for data and tools to improve management decisions. [13,24-25] The results of this conference were summarized as a “call to action” urging investigators to redouble efforts to determine optimal diagnosis and management of DCIS [24] and in turn prompted the Institute of Medicine to rank DCIS in the first quartile of topics to target comparative effectiveness research. [26]

The literature has confirmed that the patient demographic risk factors and the mammographic findings as described by radiologists according to the standardized lexicon, the Breast Imaging Reporting and Data System (BI-RADS) for mammogram feature distinctions and terminology, can predict the histology of breast cancer. [27-38] In order to investigate the breast cancer prediction problem on the aging population (the population of women over age 65) with the grades of DCIS, we conduct a tree augmented naive Bayesian network experiment trained and tested on a large clinical dataset including consecutive diagnostic mammography examinations, consequent biopsy outcomes and related cancer registry records in the population of women across all ages. The tasks of our experiment are to classify both the conventional task “Benign vs. Malignant” (“B vs. M”) and the new task “Benign/LG vs. IntG/HG/Invasive” (“B1 vs. M1”), based on mammography examination features. Note that the classifier “Malignant” in the conventional task “B vs. M” can be either DCIS or invasive cancer. Thus, although both the tasks can provide the “malignancy” (DCIS/Invasive and IntG/HG/Invasive, respectively) probabilities for the biopsy threshold setting and the biopsy decision making, the new task “B1 vs. M1” can help investigate the breast cancer prediction with respect to the grades of DCIS.

## Methodology

In general, a Bayesian network represents variables as “nodes”, which are data structures that contain an enumeration of possible values called “states” and store probabilities associated with each state, and conditional probabilities as “edges”. A naive Bayes model assumes that given the class variable, the value (“state”) of a particular feature variable is unrelated to the presence or absence of any other feature variable. Therefore in a naive Bayes model, the class variable is the “root node” and the directed arcs encode dependence relationships from the root node to the feature nodes. An important algorithm for naive Bayes model learning is to learn a tree structure to augment the edges of the naive Bayes network so as to produce a “tree augmented naive Bayesian network”. [39] Specifically, the algorithm firstly computes for each pair of feature variables the mutual information functions as the weights, secondly finds the maximum weight spanning tree and assign edge directions, and finally attaches the tree structure to the naive Bayes model to construct a tree augmented naive Bayesian network.

Figure 1 presents a typical tree augmented naive Bayesian network trained for the task “B1 vs. M1” in the experiment. The root node, entitled “Benign/LG vs. IntG/HG/Invasive”, has two states that represent the outcome of interest—“Benign/LG” or “IntG/HG/Invasive”—and stores the prior probabilities of these states. The feature nodes in the network represent demographic risk factors, BI-RADS descriptors and the BI-RADS category. And the directed arcs encode the dependence relationships among the nodes, i.e. the conditional probabilities among the variables. Note that the nature of the tree augmented naive Bayes algorithm guarantees in a tree augmented naive Bayesian network, each feature node can have no more than one dependent node besides the root node.

The tree augmented naive Bayesian network is trained and constructed on a large existing clinical dataset including consecutive diagnostic mammography examinations, consequent biopsy outcomes and related cancer registry records in the population of women across all ages. The consecutive diagnostic mammography examinations together with the patient demographic records provide the information of all the feature nodes; the root node information comes from the consequent biopsy outcomes and related cancer registry records; whereas the information of dependence relationships among the nodes is hidden in the database matching relations between the records of the consecutive diagnostic mammography examinations and the records of the consequent biopsy outcomes and related cancer registries. The model learns the probabilities within each node and discovers the arcs connecting the nodes to capture dependence relationships. As long as the Bayesian network’s predictive power is convincing in test, it can predict the posterior probability of malignancy for any new diagnostic mammography examination with patient demographic information.

## Experiment

We conduct experiment on a large clinical dataset combined by the University of Wisconsin-Madison Breast Imaging Database and the University of California San Francisco Breast Imaging Database. The UW database consists of screening and diagnostic mammography examinations at the UW Breast Imaging Center starting in October of 2005. As of 12/31/09, the database contains 41,682 mammography examinations on 24,510 patients described and recorded by the BI-RADS lexicon. The UCSF consists of 146,996 consecutive mammograms collected between 1/6/1997 and 6/29/2007. The combined dataset from UW and UCSF consists of 5607 consecutive diagnostic mammograms between 1/6/1997 and 12/27/2011 matched with following biopsy outcomes and corresponding cancer registries. 1729 cases are from UW database between 12/8/2005 and 12/27/2011 while 3878 cases are from UCSF database between 1/6/1997 and 6/29/2007.

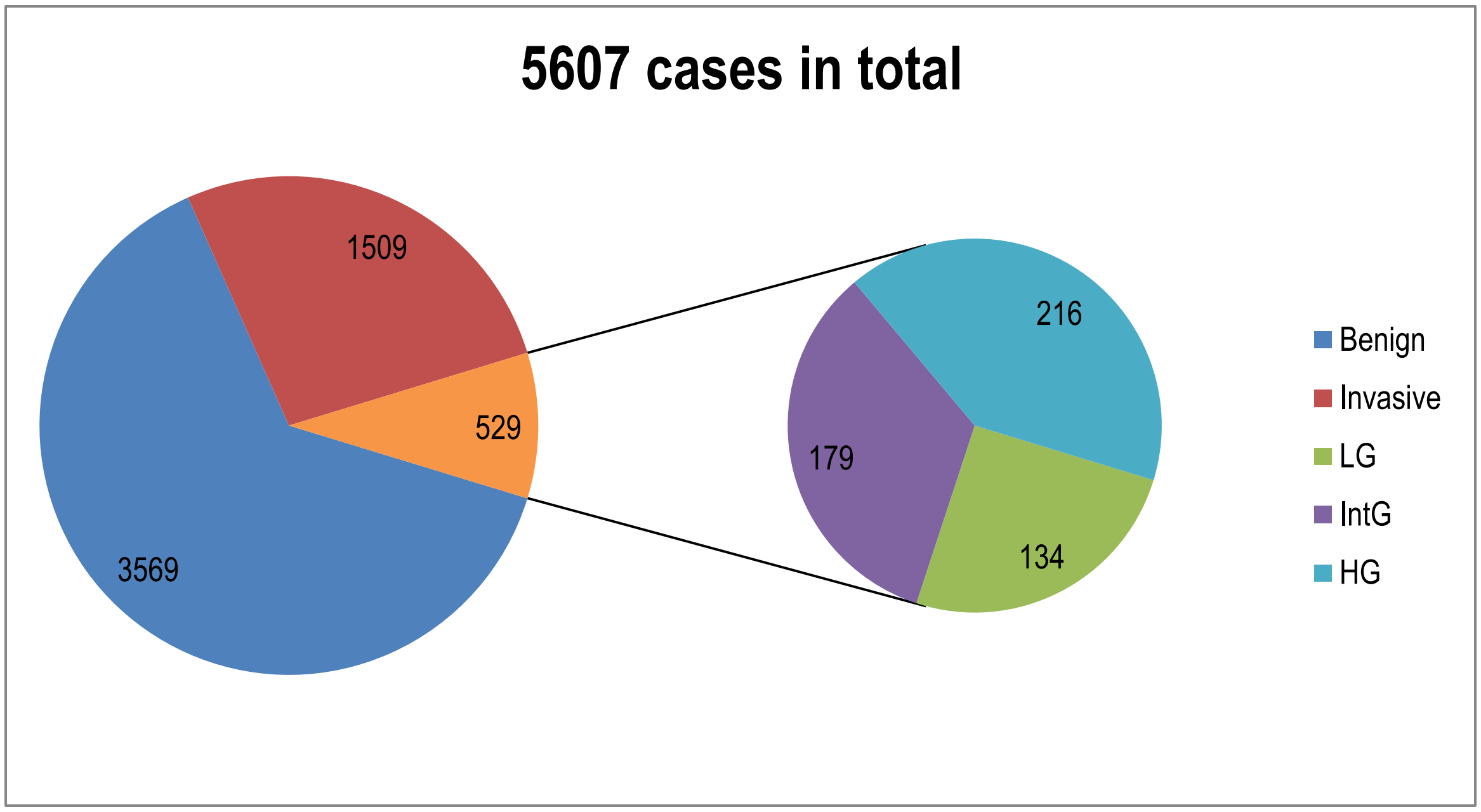

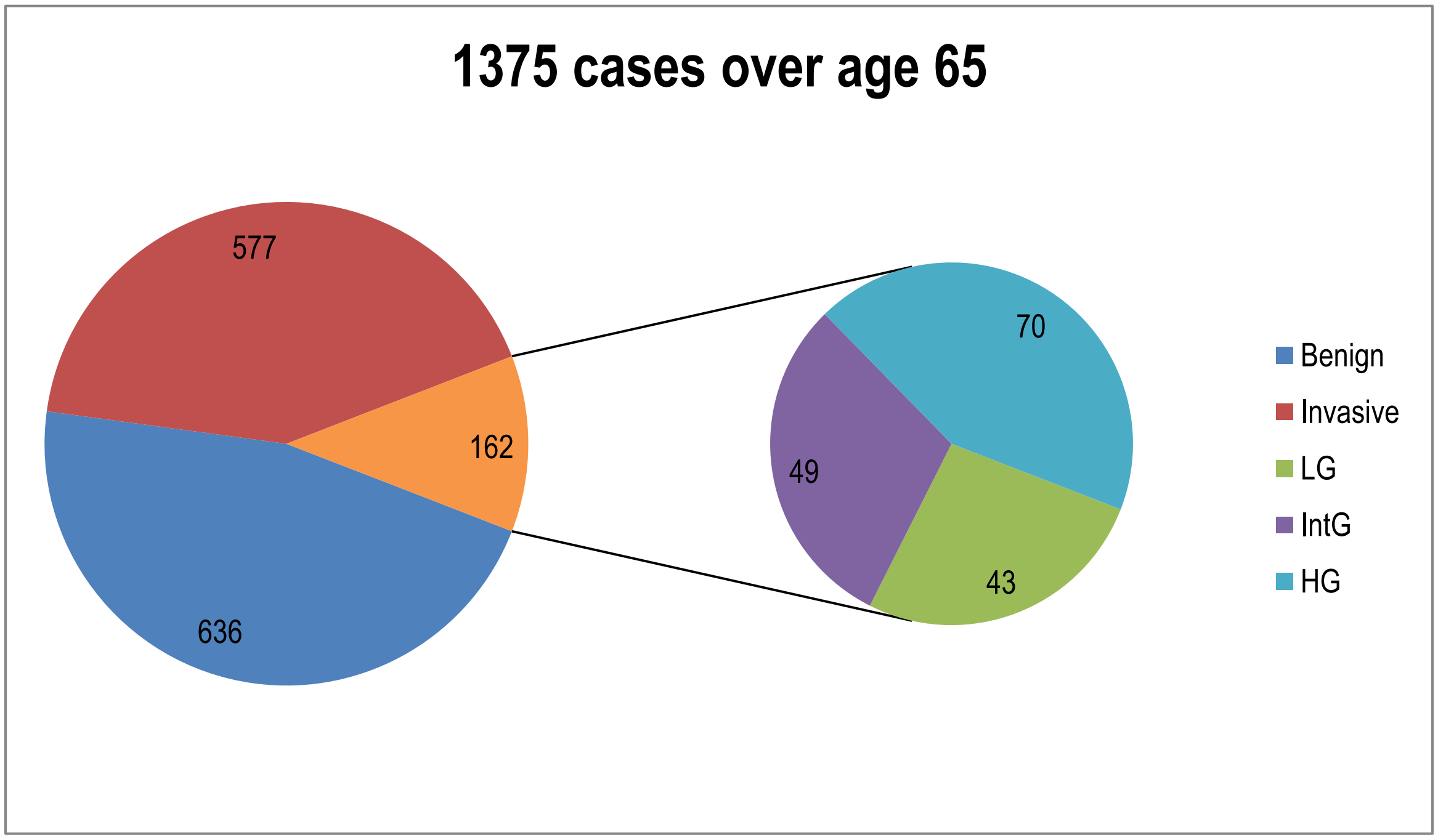

As showed in the above pie graphs, the dataset contains 3569 benign cases, 1509 invasive cases, and 529 DCIS cases in which 134 are LG, 179 are IntG, and 216 are HG; among the 1375 aging cases, the numbers of benign, invasive, DCIS, LG, IntG and HG cases are 636, 577, 43, 49 and 70, respectively. The dataset reflects a fact of diagnostic mammograms: the aging population sees a higher rate of DCIS and a much higher rate of invasive breast cancer than the average. An interesting observation is that the proportions of each of the DCIS grades seem stable.

For the experiment, we model the demographic risk factors and the mammogram features as feature nodes and the result of the biopsy outcome and/or the cancer registry as root node, follow the tree augmented naive Bayesian network algorithm to learn the probabilities and the structure from training datasets, and predict the malignancy probabilities on testing datasets.

Table 1 makes a summary of all the variables in the experiment. Especially, the variable "Age Group" is one of our demographic risk factors, with the value "Older" representing the aging population. Most of mammograms in the dataset are assigned to the BI-RADS category 0, 4 and 5. Many mammograms have "missing" values for the variable "Palpable Lump" (with value labels of "missing", "No" and "Yes"), but the "missing" values for other mammogram feature variables simply mean "no such findings".

To label the biopsy outcome for the class variable, we used the most malignant result (Invasive > DCIS) and highest grade (HG > IntG > LG) at either core biopsy or subsequent surgery during the episode of care for analysis. The "episode of care" is defined as the duration of the process to definitively determine a breast diagnosis, including a core biopsy and subsequent diagnosis. A single diagnosis may entail more than one biopsy to determine the extent of disease or to confirm a benign diagnosis. This episode of care may entail multiple biopsies over an interval of time. For our purposes, we define an episode of care as 6 months. If 2 biopsies were performed in the same breast within 6 months of each other, we considered them as in the same episode.

As the first step of the experiment, we prepare datasets for ten-fold cross validation: firstly randomly split the cases in each age group into ten equal-sized sets, each of which contains one-tenth of the benign findings, one-tenth of the invasive cancer findings and one-tenth of each of the grades of DCIS findings in that age group, along with the requirement that all records of the same patient be in the same set; secondly aggregate them into ten equal-sized folds of the whole population across all ages; and finally make ten pairs of training datasets and respective testing datasets. By using ten-fold cross validation, it is guaranteed that the cases used to train the model are never used to test the model.

The second step is the implementation of the tree augmented naive Bayes algorithm. We use Weka (Waikato Environment for Knowledge Analysis) [40] for the training and testing of the tree augmented naive Bayesian network on the ten pairs of training datasets and respective testing datasets.

Finally, from the output prediction files given by Weka in the implementation of the tree augmented naive Bayes algorithm for the training and testing of the tree augmented naive Bayesian network on the ten pairs of training datasets and respective testing datasets, we collect and aggregate the predicted malignancy probabilities on all the ten testing datasets. For the purpose of threshold analysis, for each of the 5001 possible breast biopsy thresholds, from 0.00% to 100.00%, we assume no biopsy for the cases where the predicted malignancy probabilities are below that threshold and compute the confusion matrix results in EXCEL spreadsheet with VBA macros. For the aging population, we select and aggregate the predicted malignancy probabilities of all the aging cases and compute the confusion matrix results in the same way for each of the 2001 possible breast biopsy thresholds from 0.00% to 100.00%.

In order to estimate the predictive performance of the tree augmented naive Bayesian network methodology in test, following the convention of the literature, [27-38] we construct the Receiver Operating Characteristic (ROC) curve and the Precision-Recall (PR) curve. The AUC which measures the area under the ROC curve is calculated. The AUCPR which measures the area under the PR curve is also meaningful [41-43] and is calculated as well.

## Result

For the threshold analysis, Table 2 and Table 3 make snapshots of the confusion matrix results at typical thresholds in the whole population and in the aging population, respectively for the conventional task “B vs. M” and the new task “B1 vs. M1”. From the tables we can see there are always actual malignancy cases with low malignancy probabilities predicted by the tree augmented naive Bayesian network methodology, both in the whole population and in the aging population. In fact, our EXCEL spreadsheet also shows not a few non-malignancy cases with high malignancy probabilities predicted by the methodology. But considering the number of the cases in the dataset, from a probabilistic perspective, we conclude the results are acceptable. For the task “B vs. M”, if we set the breast biopsy threshold to be 1%, no malignancy case will be missed and 22 non-malignancy cases will avoid breast biopsies. Thus 1% is the “critical threshold” and for the task “B1 vs. M1” it will save 41 non-malignancy biopsies. However, any threshold above 1% means there will be a tradeoff between avoided non-malignancy biopsies and missed malignancy biopsies. For both tasks, the conventional threshold of 2% in the whole population seems convincing, in spite of the one missed invasive aging case. In the aging population, a threshold higher than 2% would be acceptable, due to the fact that if the threshold is lifted from 2% to 7%, the number of avoided non-malignancy biopsies would rise sharply while the number of missed malignancy biopsies would rise very slowly.

Figure 2 constructs the comparable ROC curves for the conventional task “B vs. M” and the new task “B1 vs. M1”, respectively. Both the AUCs are approximately 0.84, whereas the new task yields a slightly larger one. Figure 3 constructs the comparable PR curves for the conventional task “B vs. M” and the new task “B1 vs. M1”, respectively. The AUCPR of the conventional task exceeds that of the new task by 0.01. This observation that a larger AUC does not guarantee a larger AUCPR is consistent with the literature. [41-43]

It is interesting that the PR curve can be fitted very well by a third-order polynomial curve. The third-order polynomial regression of the Precision on the Recall yields an R-square of 0.9986 with very significant regression parameters. We also find the relationship between the FPR (False Positive Rate) and the Precision can be fitted very well by a third-order polynomial curve. The third-order polynomial regression of the FPR on the Precision produces an R-square of 0.9997 with very significant regression parameters. Figure 4 presents the curve fitting and the regression result of the PR curve. And Figure 5 presents the curve fitting and the regression result of the relationship between the FPR and the Precision.

## Summary

One weakness of this experiment is that we aggregate the predicted malignancy probabilities on all the ten testing datasets to produce the threshold analysis and the ROC curve. This procedure is based on the assumption that the trainings of the tree augmented naive Bayesian network on the ten training datasets are the same. Although all the trainings follow the tree augmented naive Bayes algorithm, the differences among the ten training datasets which stem from the variance of data source, lead to different tree augmented naive Bayesian network structures and probabilities.

Another weakness is that we make threshold analysis in the aging population using the tree augmented naive Bayesian networks trained by the cases across all ages. A more solid tree augmented naive Bayesian network experiment in the aging population should use only aging cases (women over age 65) for training and testing.

In sum, we conduct a tree augmented naive Bayesian network experiment trained and tested on a large clinical dataset combined by the University of Wisconsin-Madison Breast Imaging Database and the University of California San Francisco Breast Imaging Database including consecutive diagnostic mammography examinations, consequent biopsy outcomes and related cancer registry records in the population of women across all ages. We classify the conventional task “Benign vs. Malignant” and the new task “Benign/LG vs. IntG/HG/Invasive” based on mammography examination features and patient demographic information. The aggregated predicted malignancy probabilities of our ten-fold cross validation method recommend a biopsy threshold higher than 2% for the aging population. The Receiver Operating Characteristic curves and the Precision-Recall curves by aggregating the ten-fold cross validation results are interesting.

# References

1. McCarthy EP, Burns RB, Freund KM, Ash AS, Shwartz M, Marwill SL, Moskowitz MA. Mammography use, breast cancer stage at diagnosis, and survival among older women. J Am Geriatr Soc. 2000 Oct;48(10):1226-33.
2. Randolph WM, Goodwin JS, Mahnken JD, Freeman JL. Regular mammography use is associated with elimination of age-related disparities in size and stage of breast cancer at diagnosis. Ann Intern Med. 2002 Nov 19;137(10):783-90.
3. Elmore JG, Barton MB, Moceri VM, Polk S, Arena PJ, Fletcher SW. Ten-year risk of false positive screening mammograms and clinical breast examinations. N Engl J Med. 1998 Apr 16;338(16):1089-96.
4. Mandelblatt JS, Cronin KA, Bailey S, Berry DA, de Koning HJ, Draisma G, Huang H, Lee SJ, Munsell M, Plevritis SK, Ravdin P, Schechter CB, Sigal B, Stoto MA, Stout NK, van Ravesteyn NT, Venier J, Zelen M, Feuer EJ. Effects of mammography screening under different screening schedules: model estimates of potential benefits and harms. Ann Intern Med. 2009 Nov 17;151(10):738-47.
5. Welch HG, Black WC. Overdiagnosis in cancer. J Natl Cancer Inst. 2010 May 5;102(9):605-13.
6. Projections of the total resident population by 5-year age groups and sex with special age categories: Middle Series, 2001 to 2005: Population Projections Program, Population Division, U.S. Census Bureau; 2000.
7. Ghosh K, Melton LJ, 3rd, Suman VJ, Grant CS, Sterioff S, Brandt KR, Branch C, Sellers TA, Hartmann LC. Breast biopsy utilization: a population-based study. Arch Intern Med. 2005 Jul 25;165(14):1593-8.
8. Poplack SP, Carney PA, Weiss JE, Titus-Ernstoff L, Goodrich ME, Tosteson AN. Screening mammography: costs and use of screening-related services. Radiology. 2005 Jan;234(1):79-85.
9. Day JC. Population Projections of the United States by Age, Sex, Race, and Hispanic Origin: 1995 to 2050, U.S. Bureau of the Census. Washington, DC, 1996: U.S. Government Printing Office; 1996.
10. Mokbel K, Cutuli B. Heterogeneity of ductal carcinoma in situ and its effects on management. Lancet Oncol. 2006 Sep;7(9):756-65.
11. Sanders ME, Simpson JF. Can we know what to do when DCIS is diagnosed? Oncology (Williston Park). 2011 Aug;25(9):852-6.
12. D'Orsi CJ. Imaging for the Diagnosis and Management of Ductal Carcinoma In Situ. JNCI Monographs. 2010 October 1, 2010;2010(41):214-7.
13. Allegra CJ, Aberle DR, Ganschow P, Hahn SM, Lee CN, Millon-Underwood S, Pike MC, Reed SD, Saftlas AF, Scarvalone SA, Schwartz AM, Slomski C, Yothers G, Zon R. NIH state-of-the-science conference statement: diagnosis and management of ductal carcinoma in situ (DCIS). NIH Consens State Sci Statements. 2009 Sep 24;26(2):1-27.

14. Schnitt SJ. Local Outcomes in Ductal Carcinoma In Situ Based on Patient and Tumor Characteristics. JNCI Monographs. 2010 October 1, 2010;2010(41):158-61.
15. Esserman L, Thompson I. Solving the overdiagnosis dilemma. J Natl Cancer Inst. 2010 May 5;102(9):582-3.
16. Collins LC, Tamimi RM, Baer HJ, Connolly JL, Colditz GA, Schnitt SJ. Outcome of patients with ductal carcinoma in situ untreated after diagnostic biopsy: results from the Nurses' Health Study. Cancer. 2005 May 1;103(9):1778-84.
17. Sanders ME, Schuyler PA, Dupont WD, Page DL. The natural history of low-grade ductal carcinoma in situ of the breast in women treated by biopsy only revealed over 30 years of long-term follow-up. Cancer. 2005;103(12):2481-4.
18. Li CI, Malone KE, Saltzman BS, Daling JR. Risk of invasive breast carcinoma among women diagnosed with ductal carcinoma in situ and lobular carcinoma in situ, 1988-2001. Cancer. 2006 May 15;106(10):2104-12.
19. Collins LC, Achacoso N, Nekhlyudov L, Fletcher SW, Haque R, Quesenberry CP, Jr., Puligandla B, Alshak NS, Goldstein LC, Gown AM, Schnitt SJ, Habel LA. Relationship between clinical and pathologic features of ductal carcinoma in situ and patient age: an analysis of 657 patients. The American journal of surgical pathology. 2009 Dec;33(12):1802-8.
20. Allred DC. Ductal Carcinoma In Situ: Terminology, Classification, and Natural History. JNCI Monographs. 2010 October 1, 2010;2010(41):134-8.
21. Ernster VL, Barclay J, Kerlikowske K, Grady D, Henderson C. Incidence of and treatment for ductal carcinoma in situ of the breast. JAMA. 1996 Mar 27;275(12):913-8.
22. Pisano ED. Mode of Detection and Secular Time for Ductal Carcinoma In Situ. JNCI Monographs. 2010 October 1, 2010;2010(41):142-4.
23. Virnig BA, Wang S-Y, Shamilyan T, Kane RL, Tuttle TM. Ductal Carcinoma In Situ: Risk Factors and Impact of Screening. JNCI Monographs. 2010 October 1, 2010;2010(41):113-6.
24. McCaskill-Stevens W. National Institutes of Health State-of-the-Science Conference on the Management and Diagnosis of Ductal Carcinoma In Situ: A Call to Action. JNCI Monographs. 2010 October 1, 2010;2010(41):111-2.
25. Allegra CJ, Aberle DR, Ganschow P, Hahn SM, Lee CN, Millon-Underwood S, Pike MC, Reed SD, Saftlas AF, Scarvalone SA, Schwartz AM, Slomski C, Yothers G, Zon R. National Institutes of Health State-of-the-Science Conference statement: Diagnosis and Management of Ductal Carcinoma In Situ September 22-24, 2009. J Natl Cancer Inst. 2010 Feb 3;102(3):161-9.
26. Initial National Priorities for Comparative Effectiveness Research National Academies Press; Board of Health Care Service Institute of Medicine (2009).

27. Burnside ES, Davis J, Chhatwal J, Alagoz O, Lindstrom MJ, Geller BM, Littenberg B, Shaffer KA, Kahn CE, Page CD. A Probabilistic Computer Model Developed from Clinical Data in the National Mammography Database Format to Classify Mammography Findings. Radiology. 2009;251(3):663-72.
28. Burnside ES, Rubin D, Shachter R. A Bayesian network for mammography. Proc AMIA Symp. 2000:106-10.
29. Burnside ES, Rubin DL, Fine JP, Shachter RD, Sisney GA, Leung WK. Bayesian network to predict breast cancer risk of mammographic microcalcifications and reduce number of benign biopsy results: Initial experience. Radiology. 2006;240(3):666.
30. Burnside ES, Rubin DL, Shachter RD. Using a Bayesian network to predict the probability and type of breast cancer represented by microcalcifications on mammography. Medinfo. 2004;11:13-7.
31. Burnside ES, Rubin DL, Shachter RD. Using a Bayesian network to predict the probability and type of breast cancer represented by microcalcifications on mammography. Stud Health Technol Inform. 2004;107(Pt 1):13-7.
32. Burnside ES, Davis J, Costa VS, Dutra Ide C, Kahn CE, Jr., Fine J, Page D. Knowledge discovery from structured mammography reports using inductive logic programming. AMIA Annual Symposium proceedings / AMIA Symposium. 2005:96-100.
33. Davis J, Burnside E, de Castro Dutra I, Page CD, Ramanujam N, Shavlik J, Santos Costa V. View learning for statistical relational learning: with an application to mammography. IJCAI-Proc 19th Interntl Joint Conf Artificial Intelligence. 2005:677-83.
34. Burnside ES, Ochsner JE, Fowler KJ, Fine JP, Salkowski LR, Rubin DL, Sisney GA. Use of Microcalcification Descriptors in BI-RADS 4th Edition to Stratify Risk of Malignancy. Radiology. 2007 Feb;242(2):388-95.
35. Nassif H, Woods RW, Burnside E, Ayvaci M, Shavlik J, Page CD, editors. Information extraction for clinical data mining: a mammography case study. 9th IEEE International Conference on Data Mining Workshops; 2009; Miami, FL.
36. Burnside ES, Rubin DL, Shachter RD. Improving a Bayesian Network's Ability to Predict the Probability of Malignancy of Microcalcifications on Mammography. Proc Computer Assisted Radiology and Surgery. 2004:1021-6.
37. Burnside ES, Rubin DL, Fine JP, Shachter RD, Sisney GA, Leung WK. Bayesian network to predict breast cancer risk of mammographic microcalcifications and reduce number of benign biopsy results: initial experience. Radiology. 2006 Sep;240(3):666-73.
38. Burnside ES. Bayesian networks: computer-assisted diagnosis support in radiology. Acad Radiol. 2005 Apr;12(4):422-30.
39. Friedman N, Geiger D, Goldszmidt M. Tree-augmented naïve Bayes or tree-augmented network. Mach Learn 1997;29:131–163.

40. Weka 3: Data Mining Software in Java. http://www.cs.waikato.ac.nz/ml/weka/
41. Davis J, Goadrich M. The relationship between precision-recall and ROC curves. ICML. New York ACM Press. 2006:233–240.
42. Boyd, K., Davis, J., Page, D., Santos Costa, V. Unachievable region in precision-recall space and its effect on empirical evaluation. In: Proceedings of the 29th International Conference on Machine Learning, ICML 2012, Edingburgh, Scotland.
43. Boyd K, Eng K, Page C. Area under the precision-recall curve: Point estimates and confi-dence intervals. In: Blockeel H, Kersting K, Nijssen S, Železný F, editors, Machine Learning and Knowledge Discovery in Databases, Springer Berlin Heidelberg, volume 8190 of Lecture Notes in Computer Science:451–466.

## Tables and Figures

Table 1: summary statistics of the variables in the Tree Augmented Naive Bayesian network

| **Variables** | **Instances** | | | | | | | | |
|---|---|---|---|---|---|---|---|---|---|
| **31** | **5607** | | | | | | | | |
| **Demographic** | | | | | | | | | |
| Age Group | Younger | Middle | Older | | | | | | |
| | 2091 | 2141 | 1375 | | | | | | |
| Personal History | No | Yes | | | | | | | |
| | 4697 | 910 | | | | | | | |
| Family History | None | Minor | Major | missing | | | | | |
| | 3888 | 1014 | 416 | 289 | | | | | |
| **Imaging** | | | | | | | | | |
| BIRADS Category | 0 | 1 | 2 | 3 | 4 | 5 | 7 | 8 | 9 |
| | 440 | 0 | 2 | 2 | 4513 | 650 | 0 | 0 | 0 |
| Breast Density | Predominantly Fatty | | Scattered Fibroglandular | | Heterogeneously Dense | | Extremely Dense | | missing |
| | 484 | | 2164 | | 2384 | | 574 | | 1 |
| **Mass Margin** | | | | | | | | | |
| Circumscribed | missing | present | | | | | | | |
| | 4927 | 680 | | | | | | | |
| Obscured | missing | present | | | | | | | |
| | 5195 | 412 | | | | | | | |
| Microlobulated | missing | present | | | | | | | |
| | 5561 | 46 | | | | | | | |
| Spiculated | missing | present | | | | | | | |
| | 5116 | 491 | | | | | | | |
| Indistinct | missing | present | | | | | | | |
| | 4825 | 782 | | | | | | | |
| **Mass Shape** | | | | | | | | | |
| Oval | missing | present | | | | | | | |
| | 5065 | 542 | | | | | | | |
| Round | missing | present | | | | | | | |
| | 5425 | 182 | | | | | | | |
| Lobular | missing | present | | | | | | | |
| | 5167 | 440 | | | | | | | |
| Irregular | missing | present | | | | | | | |
| | 5012 | 595 | | | | | | | |
| **Mass Density** | | | | | | | | | |
| Fat | missing | present | | | | | | | |
| | 5598 | 9 | | | | | | | |
| Low | missing | present | | | | | | | |
| | 5578 | 29 | | | | | | | |
| Equal | missing | present | | | | | | | |

| | | | |
|---|---|---|---|
| | 5201 | 406 | |
| High | missing | present | |
| | 5373 | 234 | |
| **Calcification Morphology** | | | |
| Round | missing | present | |
| | 5566 | 41 | |
| Punctate | missing | present | |
| | 5490 | 117 | |
| Amorphous | missing | present | |
| | 4950 | 657 | |
| Pleomorphic | missing | present | |
| | 4696 | 911 | |
| Fine Linear | missing | present | |
| | 5323 | 284 | |
| **Calcification Distribution** | | | |
| Diffuse | missing | present | |
| | 5434 | 173 | |
| Regional | missing | present | |
| | 5576 | 31 | |
| Clustered | missing | present | |
| | 3693 | 1914 | |
| Segmental | missing | present | |
| | 5521 | 86 | |
| Linear | missing | present | |
| | 5441 | 166 | |
| Asymmetric Density | missing | present | |
| | 5116 | 491 | |
| Architectural Distortion | missing | present | |
| | 5140 | 467 | |
| Palpable Lump | missing | No | Yes |
| | 1376 | 2560 | 1671 |

Table 2(a): typical biopsy thresholds and confusion matrix results for “B vs. M” in the whole population

| Biopsy threshold (%) | Benign biopsies | LG/IntG/HG/ Invasive biopsies | Benign biopsies avoided | LG biopsies missed | IntG biopsies missed | HG biopsies missed | Invasive biopsies missed | PPV | Sensitivity | Specificity |
|---|---|---|---|---|---|---|---|---|---|---|
| Baseline | 3569 | 2038 | 0 | 0 | 0 | 0 | 0 | 0.3635 | 1.0000 | 0.0000 |
| 0.5 | 3567 | 2038 | 2 | 0 | 0 | 0 | 0 | 0.3636 | 1.0000 | 0.0006 |
| 1.0 | 3547 | 2038 | 22 | 0 | 0 | 0 | 0 | 0.3649 | 1.0000 | 0.0062 |
| 1.5 | 3495 | 2035 | 74 | 1 | 0 | 1 | 1 | 0.3680 | 0.9985 | 0.0207 |
| 2.0 | 3437 | 2032 | 132 | 1 | 0 | 1 | 4 | 0.3715 | 0.9971 | 0.0370 |
| 2.5 | 3371 | 2028 | 198 | 4 | 0 | 1 | 5 | 0.3756 | 0.9951 | 0.0555 |
| 3.0 | 3295 | 2022 | 274 | 6 | 0 | 2 | 8 | 0.3803 | 0.9921 | 0.0768 |
| 3.5 | 3224 | 2016 | 345 | 7 | 1 | 3 | 11 | 0.3847 | 0.9892 | 0.0967 |
| 4.0 | 3140 | 2014 | 429 | 7 | 1 | 3 | 13 | 0.3908 | 0.9882 | 0.1202 |
| 5.0 | 3022 | 2005 | 547 | 8 | 2 | 5 | 18 | 0.3988 | 0.9838 | 0.1533 |

Table 2(b): typical biopsy thresholds and confusion matrix results for “B vs. M” in the aging population

| Biopsy threshold (%) | Benign biopsies | LG/IntG/HG/ Invasive biopsies | Benign biopsies avoided | LG biopsies missed | IntG biopsies missed | HG biopsies missed | Invasive biopsies missed | PPV | Sensitivity | Specificity |
|---|---|---|---|---|---|---|---|---|---|---|
| Baseline | 636 | 739 | 0 | 0 | 0 | 0 | 0 | 0.5375 | 1.0000 | 0.0000 |
| 1.5 | 636 | 739 | 0 | 0 | 0 | 0 | 0 | 0.5375 | 1.0000 | 0.0000 |
| 2.0 | 636 | 738 | 0 | 0 | 0 | 0 | 1 | 0.5371 | 0.9986 | 0.0000 |
| 3.0 | 635 | 737 | 1 | 1 | 0 | 0 | 1 | 0.5372 | 0.9973 | 0.0016 |
| 4.0 | 634 | 737 | 2 | 1 | 0 | 0 | 1 | 0.5376 | 0.9973 | 0.0031 |
| 5.0 | 629 | 737 | 7 | 1 | 0 | 0 | 1 | 0.5395 | 0.9973 | 0.0110 |
| 6.0 | 623 | 737 | 13 | 1 | 0 | 0 | 1 | 0.5419 | 0.9973 | 0.0204 |
| 7.0 | 614 | 737 | 22 | 1 | 0 | 0 | 1 | 0.5455 | 0.9973 | 0.0346 |
| 7.5 | 607 | 735 | 29 | 1 | 0 | 0 | 3 | 0.5477 | 0.9946 | 0.0456 |
| 8.0 | 604 | 735 | 32 | 1 | 0 | 0 | 3 | 0.5489 | 0.9946 | 0.0503 |
| 9.0 | 599 | 733 | 37 | 1 | 0 | 1 | 4 | 0.5503 | 0.9919 | 0.0582 |
| 10.0 | 597 | 731 | 39 | 1 | 0 | 1 | 6 | 0.5505 | 0.9892 | 0.0613 |

Table 3(a): typical biopsy thresholds and confusion matrix results for “B1 vs. M1” in the whole population

| Biopsy threshold (%) | Benign/LG biopsies | IntG/HG/ Invasive biopsies | Benign biopsies avoided | LG biopsies avoided | IntG biopsies missed | HG biopsies missed | Invasive biopsies missed | PPV | Sensitivity | Specificity |
|---|---|---|---|---|---|---|---|---|---|---|
| Baseline | 3703 | 1904 | 0 | 0 | 0 | 0 | 0 | 0.3396 | 1.0000 | 0.0000 |
| 0.5 | 3699 | 1904 | 4 | 0 | 0 | 0 | 0 | 0.3398 | 1.0000 | 0.0011 |
| 1.0 | 3662 | 1904 | 40 | 1 | 0 | 0 | 0 | 0.3421 | 1.0000 | 0.0111 |
| 1.5 | 3598 | 1902 | 104 | 1 | 0 | 0 | 2 | 0.3458 | 0.9989 | 0.0284 |
| 2.0 | 3517 | 1896 | 182 | 4 | 0 | 4 | 4 | 0.3503 | 0.9958 | 0.0502 |
| 2.5 | 3425 | 1888 | 272 | 6 | 1 | 6 | 9 | 0.3554 | 0.9916 | 0.0751 |
| 3.0 | 3343 | 1887 | 353 | 7 | 1 | 6 | 10 | 0.3608 | 0.9911 | 0.0972 |
| 3.5 | 3269 | 1881 | 425 | 9 | 2 | 7 | 14 | 0.3652 | 0.9879 | 0.1172 |
| 4.0 | 3186 | 1873 | 506 | 11 | 4 | 9 | 18 | 0.3702 | 0.9837 | 0.1396 |
| 5.0 | 2977 | 1856 | 714 | 12 | 6 | 15 | 27 | 0.3840 | 0.9748 | 0.1961 |

Table 3(b): typical biopsy thresholds and confusion matrix results for “B1 vs. M1” in the aging population

| Biopsy threshold (%) | Benign/LG biopsies | IntG/HG/ Invasive biopsies | Benign biopsies avoided | LG biopsies avoided | IntG biopsies missed | HG biopsies missed | Invasive biopsies missed | PPV | Sensitivity | Specificity |
|---|---|---|---|---|---|---|---|---|---|---|
| Baseline | 679 | 696 | 0 | 0 | 0 | 0 | 0 | 0.5062 | 1.0000 | 0.0000 |
| 1.5 | 679 | 696 | 0 | 0 | 0 | 0 | 0 | 0.5062 | 1.0000 | 0.0000 |
| 2.0 | 679 | 695 | 0 | 0 | 0 | 0 | 1 | 0.5058 | 0.9986 | 0.0000 |
| 3.0 | 677 | 695 | 1 | 1 | 0 | 0 | 1 | 0.5066 | 0.9986 | 0.0029 |
| 4.0 | 673 | 695 | 5 | 1 | 0 | 0 | 1 | 0.5080 | 0.9986 | 0.0088 |
| 4.5 | 672 | 694 | 6 | 1 | 0 | 1 | 1 | 0.5081 | 0.9971 | 0.0103 |
| 5.5 | 662 | 694 | 16 | 1 | 0 | 1 | 1 | 0.5118 | 0.9971 | 0.0250 |
| 6.0 | 655 | 693 | 23 | 1 | 0 | 1 | 2 | 0.5141 | 0.9957 | 0.0353 |
| 7.0 | 646 | 693 | 32 | 1 | 0 | 1 | 2 | 0.5176 | 0.9957 | 0.0486 |
| 8.0 | 641 | 690 | 37 | 1 | 1 | 1 | 4 | 0.5184 | 0.9914 | 0.0560 |
| 9.0 | 631 | 687 | 47 | 1 | 2 | 1 | 6 | 0.5212 | 0.9871 | 0.0707 |
| 10.0 | 624 | 686 | 54 | 1 | 2 | 1 | 7 | 0.5237 | 0.9856 | 0.0810 |

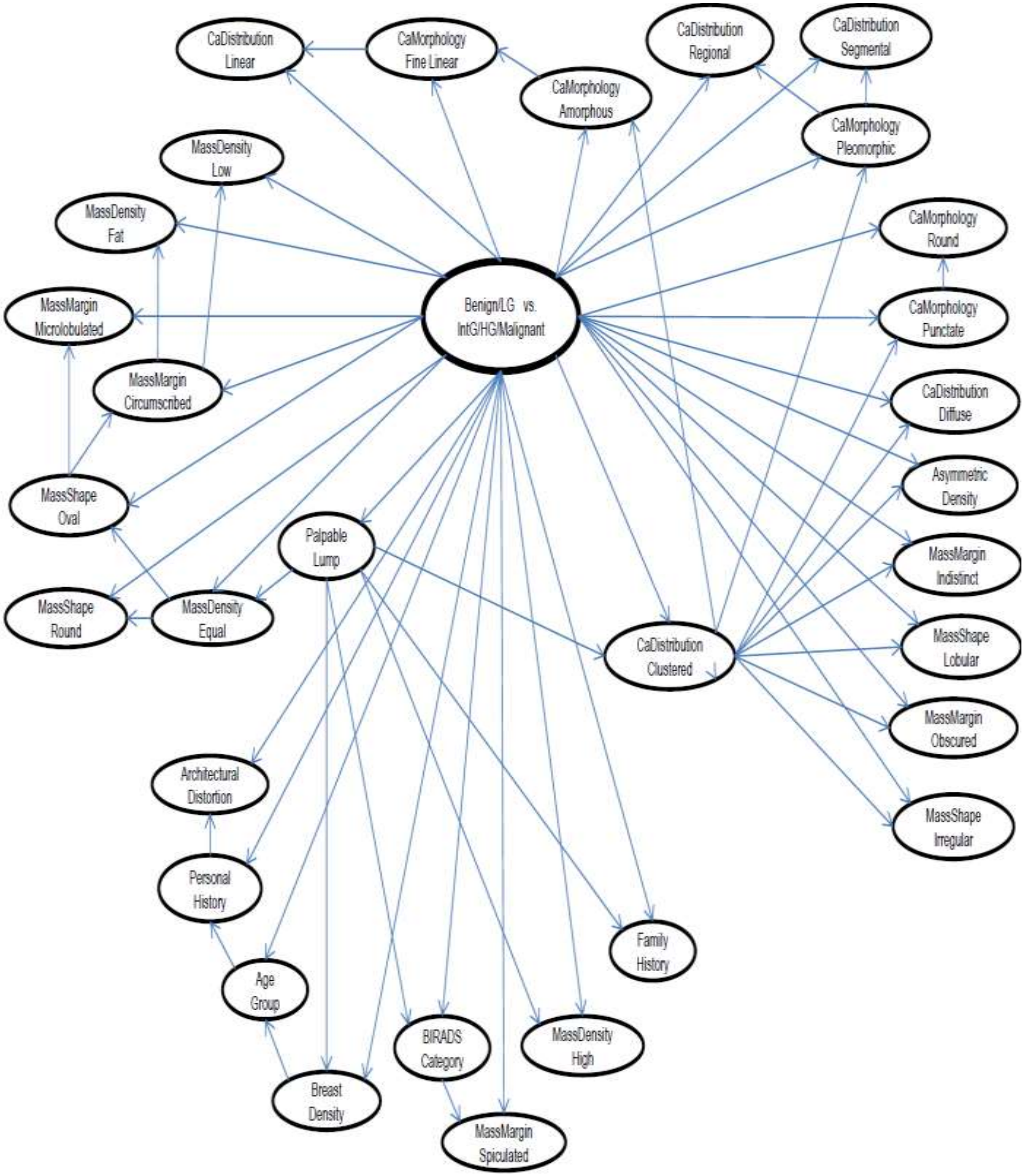

Figure 1: A typical tree augmented naive Bayesian network trained for the task “B1 vs. M1”

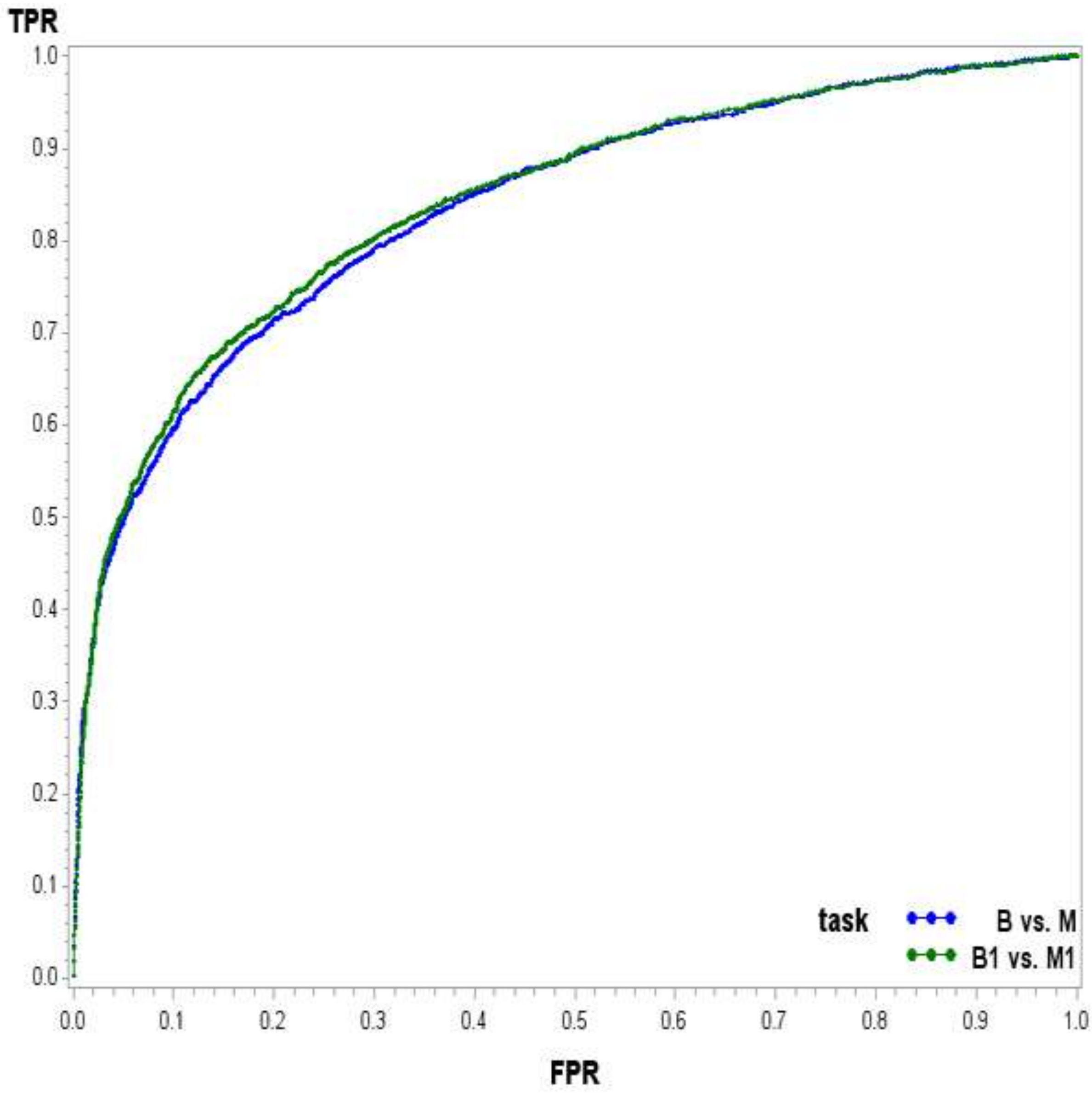

Figure 2: ROC curve for “B vs. M” (AUC = 0.836) and ROC curve for “B1 vs. M1” (AUC = 0.842)

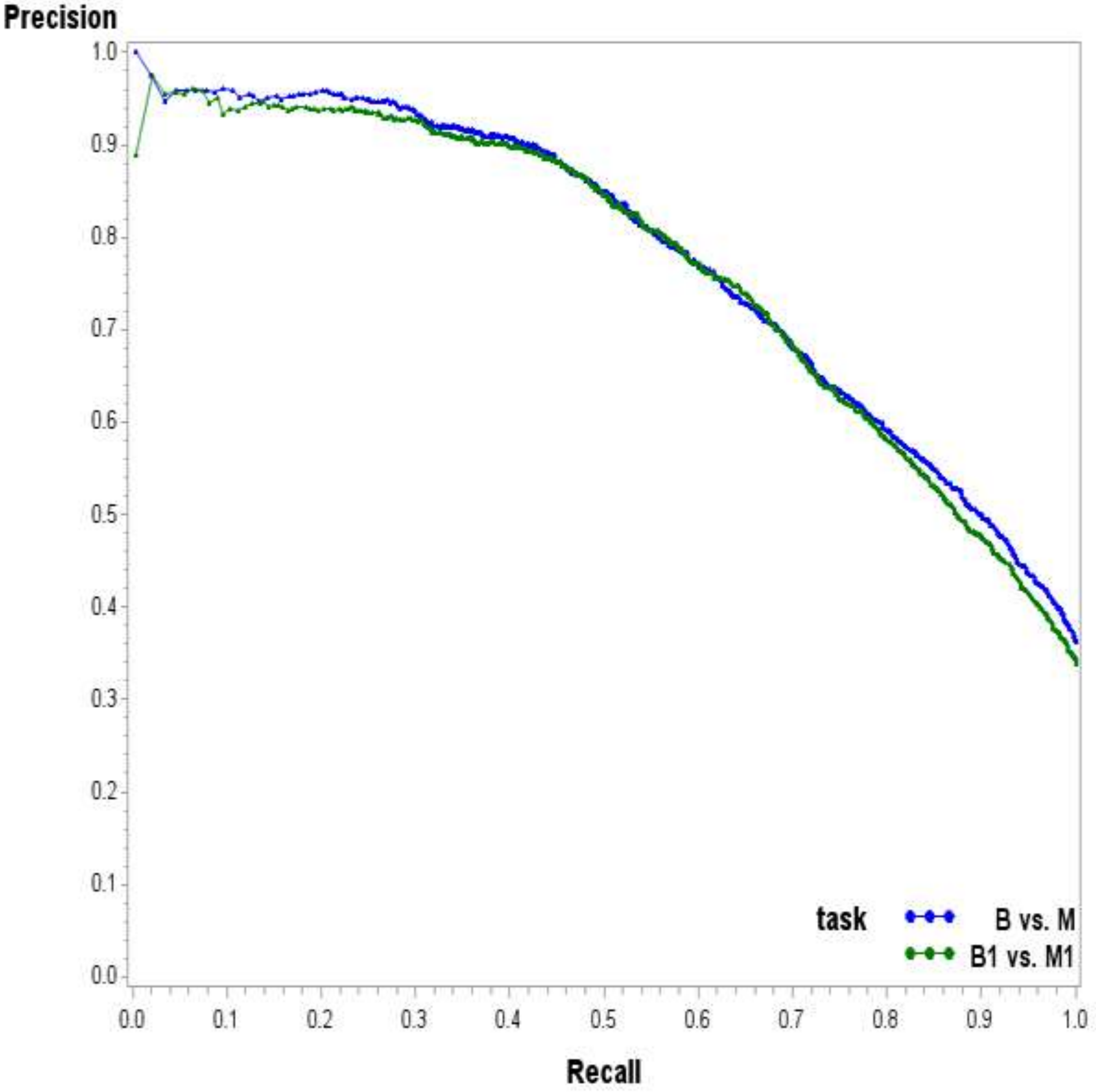

Figure 3: PR curve for “B vs. M” (AUCPR=0.781) and PR curve for “B1 vs. M1” (AUCPR=0.771)

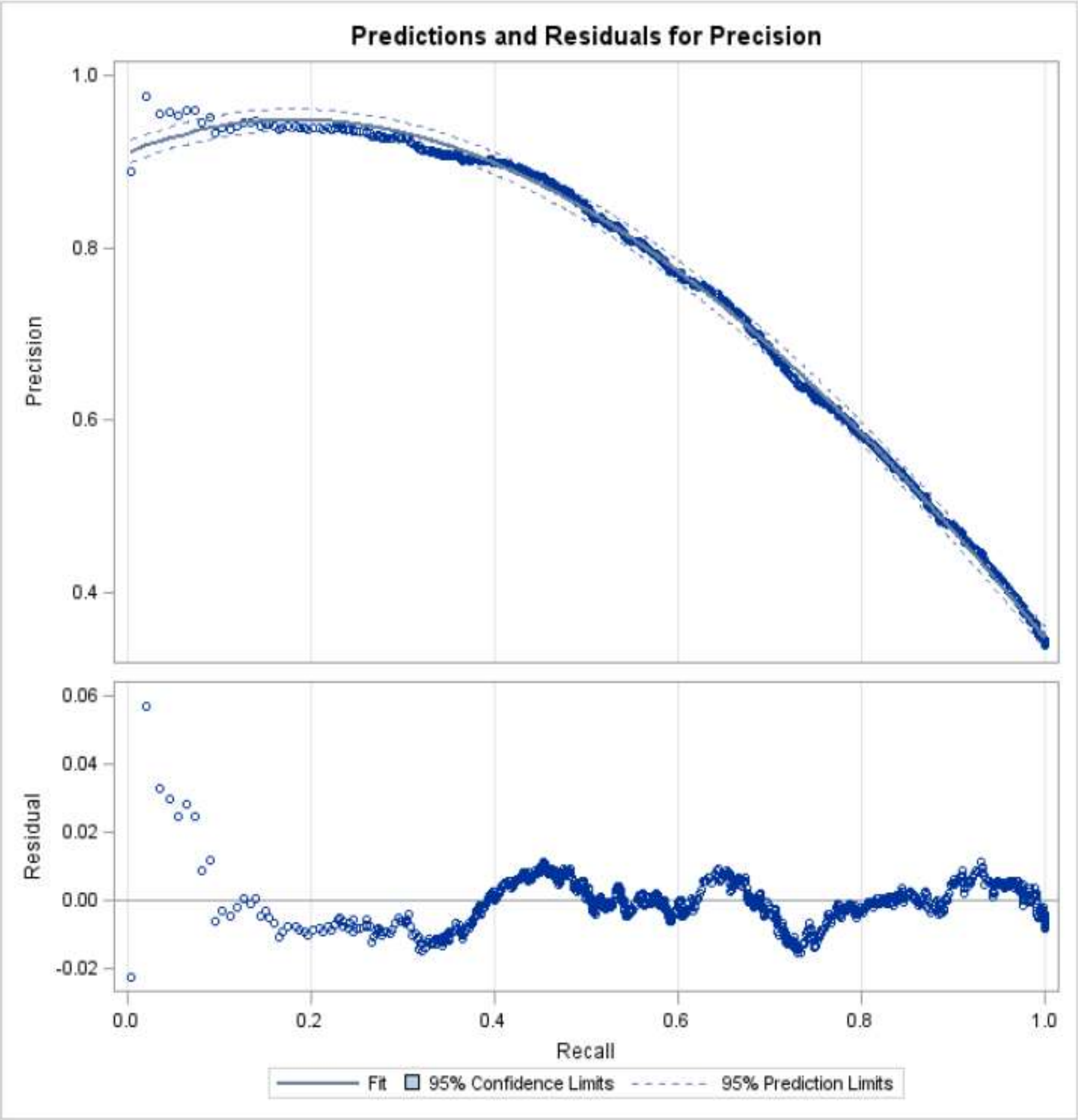

Figure 4(a): Precision-Recall curve for “B1 vs. M1” and third-order polynomial curve fitting

Dependent Variable: Precision

| Source | DF | Sum of Squares | Mean Square | F Value | Pr > F |
|---|---|---|---|---|---|
| **Model** | 3 | 150.8117664 | 50.2705888 | 1173112 | <.0001 |
| **Error** | 4997 | 0.2141331 | 0.0000429 | | |
| **Corrected Total** | 5000 | 151.0258995 | | | |

| R-Square | Coeff Var | Root MSE | Precision Mean |
|---|---|---|---|
| 0.998582 | 0.896667 | 0.006546 | 0.730056 |

| Source | DF | Type I SS | Mean Square | F Value | Pr > F |
|---|---|---|---|---|---|
| **Recall** | 1 | 140.8970567 | 140.8970567 | 3287968 | <.0001 |
| **Recall*Recall** | 1 | 9.8380581 | 9.8380581 | 229580 | <.0001 |
| **Recall*Recall*Recall** | 1 | 0.0766516 | 0.0766516 | 1788.74 | <.0001 |

| Source | DF | Type III SS | Mean Square | F Value | Pr > F |
|---|---|---|---|---|---|
| **Recall** | 1 | 0.24787548 | 0.24787548 | 5784.41 | <.0001 |
| **Recall*Recall** | 1 | 0.58365822 | 0.58365822 | 13620.2 | <.0001 |
| **Recall*Recall*Recall** | 1 | 0.07665163 | 0.07665163 | 1788.74 | <.0001 |

| Parameter | Estimate | Standard Error | t Value | Pr > \|t\| |
|---|---|---|---|---|
| **Intercept** | 0.909503979 | 0.00094941 | 957.97 | <.0001 |
| **Recall** | 0.432564855 | 0.00568750 | 76.06 | <.0001 |
| **Recall*Recall** | -1.261288606 | 0.01080743 | -116.71 | <.0001 |
| **Recall*Recall*Recall** | 0.266724594 | 0.00630652 | 42.29 | <.0001 |

Figure 4(b): Third-order polynomial regression of Precision on Recall for “B1 vs. M1”

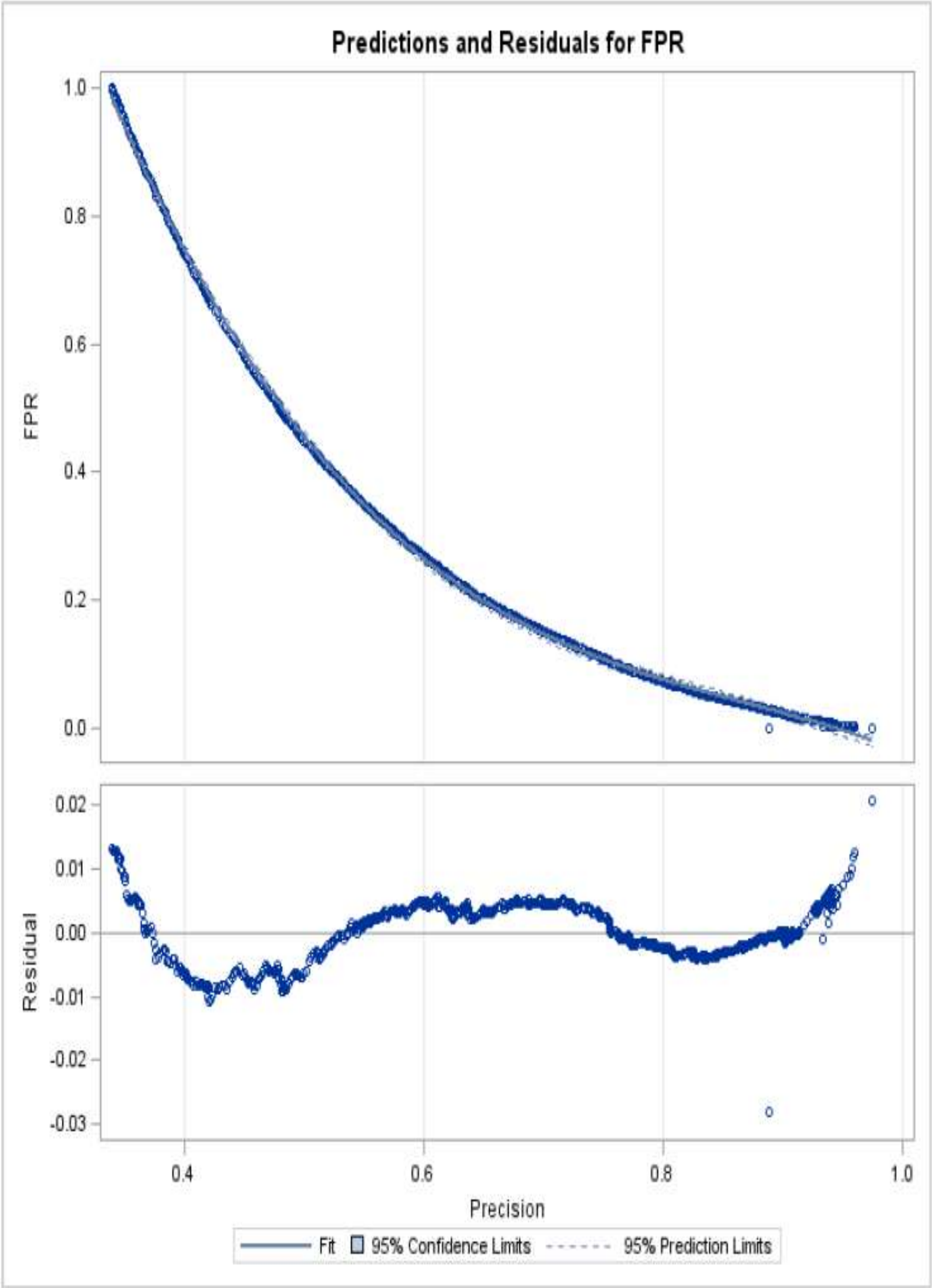

Figure 5(a): FPR-Precision relationship for “B1 vs. M1” and third-order polynomial curve fitting

Dependent Variable: FPR

| Source | DF | Sum of Squares | Mean Square | F Value | Pr > F |
|---|---|---|---|---|---|
| Model | 3 | 302.2783201 | 100.7594400 | 5486107 | <.0001 |
| Error | 4997 | 0.0917764 | 0.0000184 | | |
| Corrected Total | 5000 | 302.3700965 | | | |

| R-Square | Coeff Var | Root MSE | FPR Mean |
|---|---|---|---|
| 0.999696 | 2.183030 | 0.004286 | 0.196314 |

| Source | DF | Type I SS | Mean Square | F Value | Pr > F |
|---|---|---|---|---|---|
| Precision | 1 | 268.8134060 | 268.8134060 | 1.464E7 | <.0001 |
| Precision*Precision | 1 | 31.5595346 | 31.5595346 | 1718340 | <.0001 |
| Precision*Precision*Precision | 1 | 1.9053796 | 1.9053796 | 103743 | <.0001 |

| Source | DF | Type III SS | Mean Square | F Value | Pr > F |
|---|---|---|---|---|---|
| Precision | 1 | 6.92929507 | 6.92929507 | 377283 | <.0001 |
| Precision*Precision | 1 | 3.33434805 | 3.33434805 | 181547 | <.0001 |
| Precision*Precision*Precision | 1 | 1.90537955 | 1.90537955 | 103743 | <.0001 |

| Parameter | Estimate | Standard Error | t Value | Pr > \|t\| |
|---|---|---|---|---|
| Intercept | 3.48769185 | 0.00356822 | 977.43 | <.0001 |
| Precision | -10.89560968 | 0.01773853 | -614.23 | <.0001 |
| Precision*Precision | 11.95572794 | 0.02805959 | 426.08 | <.0001 |
| Precision*Precision*Precision | -4.58579097 | 0.01423752 | -322.09 | <.0001 |

Figure 5(b): Third-order polynomial regression of FPR on Precision for “B1 vs. M1”